\documentclass{article}
\usepackage{nips07submit_e,times,graphicx}
\usepackage{url}
\usepackage{amsmath}
\usepackage{amssymb}
\usepackage{cancel}

\newcommand{\comment}[1]  {}
\def\BE{\begin{equation}}
\def\EE{\end{equation}}
\def\BEA{\begin{eqnarray}}
\def\EEA{\end{eqnarray}}

\newtheorem{alg}{Algorithm}

\newcommand\ie{{\textsl{i.e.\,}}}
\newcommand\eg{{\textsl{e.g.\,}}}
\newcommand\etal{{\textsl{et al.\,}}}

\newcommand\vw{{\bf w}}
\newcommand\vx{{\bf x}}
\newcommand\vy{{\bf y}}

\newcommand\mI{{\bf I}}

\newcommand\mK{{\bf K}}

\newcommand\mS{{\bf S}}

\newcommand\mW{{\bf W}}

\newcommand{\ignore}[1]{}

\title{A Gaussian Belief Propagation Solver for Large Scale Support Vector Machines}

\author{
Danny Bickson \\
School of Engineering and Computer Science \\
The Hebrew University of Jerusalem \\
Givat Ram, Jerusalem, 91904 \\
Israel \\
\texttt{danny.bickson@gmail.com} \\
\And
Elad Yom-Tov \\
IBM Haifa Research Lab\\
Haifa 31905 \\
Israel \\
\texttt{yomtov@il.ibm.com} \\
\AND
Danny Dolev\\
School of Engineering and Computer Science \\
The Hebrew University of Jerusalem \\
Givat Ram, Jerusalem, 91904 \\
Israel \\
\texttt{dolev@cs.huji.ac.il} \\
}

\begin{document}

\maketitle

\begin{abstract}

Support vector machines (SVMs) are an extremely successful type of
classification and regression algorithms. Building an SVM entails
solving a constrained convex quadratic programming problem, which
is quadratic in the number of training samples. We introduce an
efficient parallel implementation of an support vector regression
solver, based on the Gaussian Belief Propagation algorithm (GaBP).

In this paper, we demonstrate that methods from the complex system
domain could be utilized for performing efficient distributed
computation. We compare the proposed algorithm to previously
proposed distributed and single-node SVM solvers. Our comparison
shows that the proposed algorithm is just as accurate as these
solvers, while being significantly faster, especially for large
datasets. We demonstrate scalability of the proposed algorithm to
up to 1,024 computing nodes and hundreds of thousands of data
points using an IBM Blue Gene supercomputer. As far as we know,
our work is the largest parallel implementation of belief
propagation ever done, demonstrating the applicability of this
algorithm for large scale distributed computing systems.

\end{abstract}

\section{Introduction}

\label{SVM classifiers}

Support-vector machines (SVMs) are a class of algorithms that
have, in recent years, exhibited superior performance compared to
other pattern classification algorithms. There are several
formulations of the SVM problem, depending on the specific
application of the SVM (e.g., classification, regression, etc.).

One of the difficulties in using SVMs is that building an SVM
requires solving a constrained quadratic programming problem,
whose size is quadratic in the number of training examples. This
fact has led to extensive research on efficient SVM solvers.
Recently, several researchers have suggested using multiple
computing nodes in order to increase the computational power
available for solving SVMs.

In this article, we introduce a distributed SVM solver based on
the Gaussian Belief Propagation (GaBP) algorithm. We improve on
the original GaBP algorithm by reducing the communication load, as
represented by the number of messages sent in each optimization
iteration, from $O(n^2)$ to $O(n)$ aggregated messages, where $n$
is the number of data points. Previously, it was known that the
GaBP algorithm is very efficient for sparse matrices. Using our
novel construction, we demonstrate that the algorithm exhibits
very good performance for dense matrices as well. We also show
that the GaBP algorithm can be used with kernels, thus making the
algorithm more powerful than what was considered previously
thought possible.

Using extensive simulation we demonstrate the applicability of our
protocol vs. the state-of-the-art existing parallel SVM solvers.
Using a Linux cluster of up to a hundred machines and the IBM Blue
Gene supercomputer we managed to solve very large data sets up to
hundreds of thousands data point, using up to 1,024 CPUs working
in parallel. Our comparison shows that the proposed algorithm is
just as accurate as these previous solvers, while being
significantly faster.

A preliminary version of this paper appeared as a poster
in~\cite{NIPS-workshop}.

\subsection{Classification Using Support Vector Machines}
We begin by formulating the SVM problem. Consider a training set:

\begin{equation}
\begin{array}{l}
D=\left\{\left(\mathbf{x}_i, y_i\right),\;\;\; i=1,\ldots,N,\;\;\;
  \mathbf{x}_i\in \Re^m, \;\;\;y_i\in \left\{-1, 1\right\}\right\}.
\end{array}
\end{equation}
The goal of the SVM is to learn a mapping from $\mathbf{x}_i$ to
$y_i$ such that the error in mapping, as measured on a new
dataset, would be minimal. SVMs learn to find the linear weight
vector that separates the two classes so that
\begin{equation}
\begin{array}{l}
y_i \left( \mathbf{x_i} \cdot \mathbf{w} + b \right) \geq 1 \;\;
for \;\; i = 1,\ldots,N.
\end{array}
\end{equation}

There may exist many hyperplanes that achieve such separation, but
SVMs find a weight vector $\mathbf{w}$ and a bias term $b$ that
maximize the margin $2 / \left\| \mathbf{w} \right\| $.
Therefore, the optimization problem that needs to be solved is
\begin{equation}
\min J_D(\mathbf{w}) = \frac{1}{2}\left\| \mathbf{w} \right\|,
\end{equation}
\begin{equation}
Subject \; to \; y_i \left( \mathbf{x_i} \cdot \mathbf{w} + b
\right) \geq 1 \;\; for \;\; i = 1,\ldots,N.
\end{equation}

Any points lying on the hyperplane $y_i \left( \mathbf{x_i} \cdot
\mathbf{w} + b \right) = 1 $ are called support vectors.

If the data cannot be separated using a linear separator, a slack
variable $\xi \geq 0$ is introduced and the constraint is relaxed
to:
\begin{equation}
y_i \left( \mathbf{x_i} \cdot \mathbf{w} + b \right) \geq 1 -
\xi_i \;\; for \;\; i = 1,\ldots,N.
\end{equation}

The optimization problem then becomes:
\begin{equation}
\min J_D(\mathbf{w}) = \frac{1}{2}\left\| \mathbf{w} \right\| + C
\sum_{i=1}^N \xi_i,
\end{equation}
\begin{equation}
subject \; to \; y_i \left( \mathbf{x_i} \cdot \mathbf{w} + b
\right) \geq 1 \;\; for \;\; i = 1,\ldots,N, \newline
\end{equation}
\begin{equation}
\xi_i \geq 0  \;\; for \;\; i = 1,\ldots,N.
\end{equation}

The weights of the linear function can be found directly or
by converting the problem into its dual optimization problem, which is usually easier to solve.

Using the notation of Vijayakumar and Wu~\cite{SVMSeq}, the dual
problem is thus:
\begin{equation}
\label{dual}
    \max \;\; L_D(h)=\sum_{i}h_i-\frac{1}{2}h'\cdot D\cdot h, \\
\end{equation}
\begin{equation}
\label{cons1}
    subject \; to \;\; 0\leq h_i \leq C, \;\; i=1,...,N,\\
\end{equation}
\begin{equation}
\label{cons2}
    \; \; \; \;\; \;\;\; \Sigma_i h_i y_i = 0.
\end{equation}

where $D$ is a matrix such that $D_{ij} = y_i y_j K
\left(\mathbf{x}_i, \mathbf{x}_j \right) $ and $K \left(\cdot, \;
\cdot\right)$ is either an inner product of the samples or a
function of these samples. In the latter case, this function is
known as the kernel function, which can be any function that
complies with the Mercer conditions~\cite{SS2002}. For example,
these may be polynomial functions, radial-basis (Gaussian)
functions, or hyperbolic tangents. If the data is not separable,
$C$ is a tradeoff between maximizing the margin and reducing the
number of misclassifications.

        The classification of a new data point is then computed using the following equation:
\begin{equation}
         \left( x \right) = sign \left( \sum_{i\in SV} h_i y_i K \left( x_i, x \right) + b \right)
\end{equation}

\subsection{Kernel Ridge Regression problem} Kernel Ridge Regression (KRR) implements a regularized form of
the least squares method useful for both regression and
classification. The non-linear version of KRR is similar to the
Support-Vector Machine (SVM) problem. However, in the latter,
special emphasis is given to points close to the decision
boundary, which is not provided by the cost function used by KRR.

Given training data \[ \mathcal{D} = \{\vx_i, y_i\}_{i=1}^{l},
\mbox{   }\vx_i \in R^{d} \mbox{
  }, y_i \in R \] the KRR algorithm determines the parameter
  vector $\vw \in R^d$ of a non-linear model (using the ``kernel trick''),
 via minimization of the following objective function:~\cite{SVM}:
\[ \min \lambda ||\vw||^2 + \sum_{i=1}^l (y_i - \vw^T\Phi(\vx_i))^2\]
where $\lambda$ is a tradeoff parameter between the two terms of
the optimization function, and $\Phi(\dot)$ is a (possible
non-linear) mapping of the training patterns.

One can show that the dual form of this optimization problem is
given by:
\begin{equation}
\label{cost} \max W(\alpha) = \vy^Ta + 1/4\lambda \alpha^T \mK
\alpha -
 1/4\alpha^T\alpha
\end{equation}
where $\mK$ is a matrix whose $(i,j)$-th entry is the kernel
function $\mK_{i,j}= \Phi(\vx_i)^T \Phi(\vx_j)$.

The optimal solution to this optimization problem is: \[ \alpha =
2\lambda(\mK+\lambda \mI)^{-1}\vy
\]

The corresponding prediction function is given by:
\[ f(\vx) = \vw^T \Phi(\vx) = \vy^T(\mK+\lambda \mI)^{-1}\mK(\vx_i,\vx). \]

The underlying assumption used is that the kernel matrices are
invertible.
\subsection{Previous Approaches for Solving Parallel SVMs}
        There are several main methods for finding a solution to an SVM problem
        on a single-node computer. (See Chapter 10 of~\cite{SS2002}) for a taxonomy
        of such methods.) However, since solving an SVM is quadratic in time and
        cubic in memory, these methods encounter difficulty when scaling to datasets
        that have many examples and support vectors. The latter two are not synonymous.
        A large dataset with many repeated examples might be solved using sub-sampling
        approaches, while a highly non-separable dataset with many support vectors will
         require an altogether different solution strategy.
        The literature covers several attempts at solving SVMs in parallel,
        which allow for greater computational power and larger memory size. In Collobert et al.~\cite{CBB2002}
        the SVM solver is parallelized by training multiple SVMs, each on a subset
        of the training data, and aggregating the resulting classifiers into a single
        classifier. The training data is then redistributed to the classifiers according
         their performance and the process is iterated until convergence is reached.
         The need to re-divide the data among the SVM classifiers means that the data
         must be moved between nodes several times; this rules out the use of an
         approach where bandwidth is a concern.
        A more low-level approach is taken by Zanghirati et al.~\cite{ZZ2003}, where the quadratic
         optimization problem is divided into smaller quadratic programs (similar to the
         Active Set methods), each of which is solved on a different node. The results
         are aggregated and the process is repeated until convergence. The performance
         of this method has a strong dependence on the caching architecture of the cluster.
        Graf et al.~\cite{GCBD2004} partition the data and solve an SVM for each partition. The support
        vectors from each pair of classifiers are then aggregated into a new training set
        for which an SVM is solved. The process continues until a single classifier remains.
         The aggregation process can be iterated, using the support vectors of the final
         classifier in the previous iteration to seed the new classifiers. One problem
         with this approach is that the data must be repeatedly shared between nodes,
         meaning that once again the goal of data distribution cannot be attained. The
         second problem, which might be more severe, is that the number of possible
         support vectors is restricted by the capacity of a single SVM solver.
         Yom Tov~\cite{SVMSeqY} proposed modifying the sequential algorithm
        developed in~\cite{SVMSeq} to batch mode. In this way, the complete
        kernel matrix is held in distributed memory and the Lagrange multipliers
         are computed iteratively. This method has the advantage that it can
         efficiently solve difficult SVM problems that have many
          support vectors to their solution. Based on that work, we show in this
          paper how an SVM solution can be obtained by adapting a Gaussian Belief
          Propagation algorithm to the solution of the algorithm proposed in~\cite{SVMSeq}.

Recently, Hazan \etal proposed an iterative algorithm for parallel
decomposition based on Fenchel Duality~\cite{Hazan}. Zanni \etal
proposes a decomposition method for computing SVM in
parallel~\cite{Zanni}. We compare our run time results to both
systems in Section~\ref{exp_results}.

\section{Gaussian Belief Propagation}

\label{gabp}
In this section we present our novel contribution - a Gaussian
Belief Propagation solver for distributed computation of the SVM
problem~\cite{ISIT1,ISIT2,Allerton}.

Following, we provide a step-by-step derivation of our GaBP solver
from our proposed cost function. As stated in the previous
section, our aim is to find $\vx^{\ast}$, a solution to the
quadratic cost function given in eq. (~\ref{dual}). Using linear
algebra notation, we can rewrite the same cost function:
\mbox{$\min E(\vx)\triangleq\vx^{T}\mW\vx/2-\vy^{T}\vx$}.

As the matrix $\mW$ is symmetric \footnote{an extension to
non-symmetric matrices is discussed in ~\cite{ISIT2}. For
simplicity of arguments, we handle the symmetric case in this
paper} (\eg, \mbox{$\mW=\mS^{T}\mS$}, the derivative of the
quadratic form with respect to the vector $\vx$ is given by
$E'(\vx)=\mW\vx-\vy$.

Thus, equating $E'(\vx)=\mathbf{0}$ gives the global minimum
$\vx^{\ast}$ of this convex function, which is the desired
solution
 $\vx^{\ast}=\mW^{-1}\vy$.

Now, one can define the following \emph{jointly Gaussian}
distribution \BE\label{eq_G}
       p(\vx)\triangleq\mathcal{Z}^{-1}\exp{\big(-E(\vx)\big)}=\mathcal{Z}^{-1}\exp{(-\vx^{T}\mW\vx/2+\vy^{T}\vx)},\EE where
$\mathcal{Z}$ is a distribution normalization factor. Defining the
vector \mbox{$\mathbf{\mu}\triangleq\mW^{-1}\vy$},    one gets the
form \BEA
p(\vx)&=&\mathcal{Z}^{-1}\exp{(\mathbf{\mu}^{T}\mW\mathbf{\mu}/2)}\nonumber\\&\times&
\exp{(-\vx^{T}\mW\vx/2+\mathbf{\mu}^{T}\mW\vx-\mathbf{\mu}^{T}\mW\mathbf{\mu}/2)}
\nonumber\\&=&\mathcal{\zeta}^{-1}\exp{\big(-\frac{1}{2}(\vx-\mathbf{\mu})^{T}\mW(\vx-\mathbf{\mu})\big)}
\nonumber\\&=&\mathcal{N}(\mathbf{\mu},\mW^{-1}),\EEA where the
new normalization factor
{$\mathcal{\zeta}\triangleq\mathcal{Z}\exp{(-\mathbf{\mu}^{T}\mW\mathbf{\mu}/2)}$}.
To summarize to this point, the target solution
$\vx^{\ast}=\mW^{-1}\vy$ is equal to
$\mathbf{\mu}\triangleq\mW^{-1}\vy$, which is the mean vector of
the distribution $p(\vx)$~, as defined in eq. (\ref{eq_G}).

The formulation above allows us to shift the rating problem from
an algebraic to a probabilistic domain. Instead of solving a
deterministic vector-matrix linear equation, we now solve an
inference problem in a graphical model describing a certain
Gaussian distribution function. Given the Peer-to-Peer graph $\mW$
and the prior vector $\vy$, one knows how to write explicitly
$p(\vx)$~(\ref{eq_G}) and the corresponding graph $\mathcal{G}$
with edge potentials (compatibility functions) $\psi_{ij}$ and
self-potentials (`evidence') $\phi_{i}$. These graph potentials
are determined according to the following pairwise factorization
of the Gaussian distribution $p(\vx)$~(\ref{eq_G})\BE
        p(\vx)\propto\prod_{i=1}^{K}\phi_{i}(x_{i})\prod_{\{i,j\}}\psi_{ij}(x_{i},x_{j}),\EE
        resulting in \mbox{$\psi_{ij}(x_{i},x_{j})\triangleq \exp(-x_{i}A_{ij}x_{j})$} and
        \mbox{$\phi_{i}(x_{i})=\exp\big(b_{i}x_{i}-A_{ii}x_{i}^{2}/2\big)$}.
        The set of edges $\{i,j\}$ corresponds to the set of
        network edges $\mW$. Hence, we would like to calculate the
marginal densities, which must also be Gaussian,
\[ p(x_{i})\sim\mathcal{N}(\mu_{i}=\{\mW^{-1}\vy\}_{i},P_{i}^{-1}=\{\mW^{-1}\}_{ii}), \]
where $\mu_{i}$ and $P_{i}$ are the marginal mean and inverse
variance (a.k.a. precision), respectively. Recall that, according
to our previous argumentation, the inferred mean $\mu_{i}$ is
identical to the desired solution $x_{i}^{\ast}$.

The move to the probabilistic domain calls for the utilization of
BP as an efficient inference engine. The sum-product rule of BP
for \emph{continuous} variables, required in our case, is given
by~\cite{Weiss} \begin{equation}\label{eq_contBP}
    m_{ij}(x_j) = \alpha \int_{x_i} \psi_{ij}(x_i,x_j) \phi_{i}(x_i)
\prod_{k \in \mathcal{N}(i)\setminus j} m_{ki}(x_i) dx_{i},
\end{equation} where $m_{ij}(x_j)$ is the message sent from node $i$ to node $j$ over their shared edge on the graph, scalar $\alpha$  is a normalization constant and the set $\mathcal{N}(i)\backslash j$ denotes all the nodes neighboring
$x_{i}$, except $x_{j}$. The marginals are computed according to
the product rule~\cite{Weiss} \BE\label{eq_product}
p(x_{i})=\alpha
\phi_{i}(x_{i})\prod_{k\in\mathcal{N}(i)}m_{ki}(x_{i}). \EE

GaBP is a special case of continuous BP where the underlying
distribution is Gaussian. In~\cite{Allerton} we show how to derive
the GaBP update rules by substituting Gaussian distributions in
the continuous BP equations. The output of this derivation is
update rules that are computed locally by each node. The
GaBP-based implementation of the Peer-to-Peer rating algorithm is
summarized in Table~1\comment{\ref{tab_summary}}.


\begin{table}[h!]
\begin{alg}\label{alg_GaBP}\end{alg}\centerline{
\begin{tabular}{|lll|}
  \hline&&\\
  \texttt{1.} & \emph{\texttt{Initialize:}} & $\checkmark$\quad\texttt{Set the neighborhood} $\textrm{N}(i)$ \texttt{to include}\\&&\quad\quad$\forall k\neq i \exists A_{ki}\neq0$.\\&& $\checkmark$\quad\texttt{Set the scalar fixes}\\&&$\quad\quad P_{ii}=A_{ii}$ \texttt{and} $\mu_{ii}=b_{i}/A_{ii}$, $\forall i$.\\
  && $\checkmark$\quad\texttt{Set the initial $\textrm{N}(i)\ni k\rightarrow i$ scalar messages}\\&&\quad\quad $P_{ki}=0$ \texttt{and} $\mu_{ki}=0$.\\&& $\checkmark$\quad \texttt{Set a convergence threshold} $\epsilon$.\\
  {\texttt{2.}} & {\emph{\texttt{Iterate:}} }&  $\checkmark$\quad\texttt{Propagate the $\textrm{N}(i)\ni k\rightarrow i$ messages}\\&&$\quad\quad P_{ki}$ \texttt{and} $\mu_{ki}$, $\forall i$ \texttt{(under certain scheduling)}.\\&&
  $\checkmark$\quad\texttt{Compute the} $\textrm{N}(j)\ni i\rightarrow j$ \texttt{scalar messages} \\&& $\quad\quad P_{ij} = -A_{ij}^{2}/\big(P_{ii}+\sum_{{k}\in\textrm{N}(i) \backslash j}
P_{ki}\big)$,\\&& $\quad\quad\mu_{ij} =
\big(P_{ii}\mu_{ii}+\sum_{k \in \textrm{N}(i) \backslash j}
P_{ki}\mu_{ki}\big)/A_{ij}$.\\
  {\texttt{3.}} & {\emph{\texttt{Check:}}} & $\checkmark$\quad\texttt{If the messages} $P_{ij}$ \texttt{and} $\mu_{ij}$ \texttt{did not}\\&&\quad\quad\texttt{converge (w.r.t. $\epsilon$),} \texttt{return to
    Step 2.}\\&&$\checkmark$\quad\texttt{Else, continue to Step 4.}\\
  {\texttt{4.}} & {\emph{\texttt{Infer:}}} & $\checkmark$\quad\texttt{Compute the marginal means}\\&&{\quad\quad$\mu_{i}=\big(P_{ii}\mu_{ii}+\sum_{k \in
\textrm{N}(i)}P_{ki}\mu_{ki}\big)/\big(P_{ii}+\sum_{{k}\in\textrm{N}(i)}
P_{ki}\big)$, $\forall i$.}\\&&$(\checkmark$\quad\texttt{Optionally compute the marginal precisions}\\&&\quad\quad$P_{i}=P_{ii}+\sum_{k\in\textrm{N}(i)}P_{ki}\quad)$\\
  {\texttt{5.}} & {\emph{\texttt{Solve:}}} & $\checkmark$\quad\texttt{Find the solution}\\&& {$\quad\quad x_{i}^{\ast}=\mu_{i}$, $\forall i$.}\\&&\\\hline
\end{tabular}
}
\end{table}

Algorithm 1 can be easily executed distributively. Each node $i$
receives as an input the $i$'th row (or column) of the matrix
$\mW$ and the scalar $b_{i}$. In each iteration, a message
containing two reals, $\mu_{ij}$ and $P_{ij}$, is sent to every
neighboring node through their mutual edge, corresponding to a
non-zero $A_{ij}$ entry.
\ignore{ For a dense matrix $\mW$, each of the $K$ nodes sends a
unique message to every other node on the fully-connected graph,
which results in a total of $K^2$ messages per iteration round.}

\paragraph{Convergence.}
If it converges, GaBP is known to result in exact
inference~\cite{Weiss}. In contrast to conventional iterative
methods for the solution of systems of linear equations, for GaBP,
determining the exact region of convergence and convergence rate
remain open research problems. All that is known is a sufficient
(but not necessary) condition~\cite{WS,JMLR}
stating that GaBP converges when the spectral radius satisfies
\mbox{$\rho(|\mI_{K}-\mW|)<1$}. A stricter sufficient
condition~\cite{Weiss}, actually proved earlier, determines that
the matrix $\mW$ must be diagonally dominant (\ie,
$|W_{ii}|>\sum_{j\neq i}|W_{ij}| , \forall i$) in order for GaBP
to converge.

\ignore{ The number of messages passed on the graph can be reduced
significantly, down to only $K$ messages per round.\footnote{By
using a similar construction to Bickson \etal~\cite{BroadcastBP}.}
Instead of sending a message composed of the pair $\mu_{ij}$ and
$P_{ij}$, a node can broadcast the aggregated sums \BEA
\tilde{P}_{i}&=&P_{ii}+\sum_{{k}\in\mathcal{N}(i)}
P_{ki},\\\tilde{\mu}_{i}&=&P_{i}^{-1}(P_{ii}\mu_{ii}+\sum_{k \in
\mathcal{N}(i)} P_{ki}\mu_{ki}). \EEA Now, each node locally
retrieves the $P_{i\backslash j}$~(\ref{eq_prec}) and
$\mu_{i\backslash j}$~(\ref{eq_mean}) from the sums by means of a
subtraction \BEA P_{i\backslash
j}&=&\tilde{P}_{i}-P_{ji},\\\mu_{i\backslash
j}&=&\tilde{\mu}_{i}-P_{i \backslash j}^{-1}P_{ji}\mu_{ji}.\EEA
The rest of the algorithm remains the same. }

\paragraph{Efficiency.}

The local computation at a node at each round is fairly minimal.
Each node $i$ computes locally two scalar values $\mu_{ij}$,
$P_{ij}$ for each neighbor $j \in N(i)$. Convergence time is
dependent on both the inputs and the network topology. Empirical
results are provided in Section~\ref{exp_results}.

\section{Proposed Solution of SVM Solver Based on GaBP}

For our proposed solution, we take the exponent of dual SVM
formulation given in equation (\ref{dual}) and solve $\max
\exp(L_D( h))$. Since $\exp(L_D( h))$ is convex, the solution of
$\max \exp(L_D( h))$ is a global maximum that also satisfies $\max
L_D(h)$ since the matrix $D$ is symmetric and positive definite.
Now we can relate to the new problem formulation as a probability
density function, which is in itself Gaussian:
\[ p(h) \propto \exp(-\frac{1}{2}h'Dh + h'1), \] where $1$ is a
vector of $(1,1,\cdots,1)$ and find the assignment of $\hat{h} =
\arg \max p(h)$. It is known~\cite{JMLR} that in Gaussian models
finding the MAP assignment is equivalent to solving the inference
problem. To solve the inference problem, namely computing the
marginals $\hat{h}$, we propose using the GaBP algorithm, which is
a distributed message passing algorithm. We take the computed
$\hat{h}$ as the Lagrange multiplier weights of the support
vectors of the original SVM data points and apply a threshold for
choosing data points with non-zero weight as support vectors.

Note that using this formulation we ignore the remaining
constraints~\ref{cons1},~\ref{cons2}. In other words we do not
solve the SVM problem, but the kernel ridge regression problem.
Nevertheless, empirical results presented in
Section~\ref{emp_results} show that we achieve very good
classification vs. state-of-the-art SVM solvers.

Finally, following \cite{SVMSeq}, we remove the explicit bias term
$b$ and instead add another dimension to the pattern vector
$\mathbf{x}_i$ such that $\mathbf{\acute{x_i}} = \left(x_1, x_2,
\ldots, x_N, \lambda \right)$, where $\lambda$ is a scalar
constant. The modified weight vector, which incorporates the bias
term, is written as $\mathbf{\acute{w}} = \left( w_1, w_2, \ldots,
w_N,b/\lambda \right)$. However, this modification causes a change
to the optimized margin. Vijayakumar and Wu \cite{SVMSeq} discuss
the effect of this modification and reach the conclusion that
``setting the augmenting term to zero (equivalent to neglecting
the bias term) in high dimensional kernels gives satisfactory
results on real world data''. We did not completely neglect the
bias term and in our experiments, which used the Radial Basis
Kernel, set it to $1/N$, as proposed in \cite{SVMSeqY}.

\subsection{GaBP Algorithm Convergence}
\label{diagonal} In order to force the algorithm to converge, we
artificially weight the main diagonal of the kernel matrix $D$ to
make it diagonally dominant. Section~\ref{experimental} outlines
our empirical results showing that this modification did not
significantly affect the error in classifications on all tested
data sets.

A partial justification for weighting the main diagonal is found in~\cite{SVM}.
In the 2-Norm soft margin formulation of the SVM problem, the sum of squared slack
variables is minimized:
\[ \min_{\xi,w,b} \|\mathbf{w} \|_2^2 + C \Sigma_i \mathbf{\xi}_i^2 \]
\[ s.t. \mbox{     } y_i(\mathbf{w} \cdot \mathbf{x}_i + b) \ge 1 - \xi_i \]
The dual problem is derived:
\[ W(h) = \Sigma_{i} h_i - \frac{1}{2}\Sigma_{i,j} y_i y_j h_i h_j(\mathbf{x}_i \cdot \mathbf{x}_j +
                                                                      \frac{1}{C} \delta_{ij}), \]
where $\delta_{ij}$ is the Kronecker $\delta$ defined to be 1 when
$i=j$, and zero elsewhere. It is shown that the only change
relative to the 1-Norm soft margin SVM is the addition of $1/C$ to
the diagonal of the inner product matrix associated with the
training set. This has the effect of adding $1/C$ to the
eigenvalues, rendering the kernel matrix (and thus the GaBP
problem) better conditioned~\cite{SVM}.

\subsection{Convergence in Asynchronous Settings}
One of the desired properties of a large scale algorithm
is that it should converge in asynchronous settings as well as in synchronous settings. This
is because in a large-scale communication network, clocks are not
synchronized accurately and some nodes may be slower than others,
while some nodes experience longer communication delays.

Recent work by Koller et. al~\cite{RBP} defines conditions for the
convergence of belief propagation. This work defines a distance
metric on the space of BP messages; if this metric forms a
max-norm construction, the BP algorithm converges under some
assumptions. Using experiments on various network sizes, of up to
a sparse matrix of one million over one million nodes, the
algorithm converged asynchronously in all cases where it converged
in synchronous settings. Furthermore, as noted in~\cite{RBP}, in
asynchronous settings the algorithm converges faster as compared
to synchronous settings.


\section{Algorithm Optimization}

Instead of sending a message composed of the pair of $\mu_{ij}$
and $P_{ij}$, a node broadcasts aggregated sums, and consequently
each node can retrieve locally the $P_{i\backslash j}$~ and
$\mu_{i\backslash j}$ from the sums by means of a subtraction:

\begin{table}[!h!]
\begin{alg}\label{alg_GaBP_Broadcast}\end{alg}
\centerline{
\begin{tabular}{|lll|}
  \hline&&\\
  \texttt{1.} & \emph{\texttt{Initialize:}} & $\checkmark$\quad\texttt{Set the neighborhood} $\textrm{N}(i)$ \texttt{to include}\\&&\quad\quad$\forall k\neq i \exists A_{ki}\neq0$.\\&& $\checkmark$\quad\texttt{Set the scalar fixes}\\&&$\quad\quad P_{ii}=A_{ii}$ \texttt{and} $\mu_{ii}=b_{i}/A_{ii}$, $\forall i$.\\
  && $\checkmark$\quad\texttt{Set the initial $i\rightarrow\textrm{N}(i)$ broadcast messages}\\&&\quad\quad $\tilde{P_{i}}=0$ \texttt{and} $\tilde{\mu}_{i}=0$.\\&&$\checkmark$\quad\texttt{Set the initial $\textrm{N}(i)\ni k\rightarrow i$ internal scalars}\\&&\quad\quad $P_{ki}=0$ \texttt{and} $\mu_{ki}=0$.\\&& $\checkmark$\quad \texttt{Set a convergence threshold} $\epsilon$.\\
  {\texttt{2.}} & {\emph{\texttt{Iterate:}} }&  $\checkmark$\quad\texttt{Broadcast the aggregated sum messages}\\&&$\quad\quad \tilde{P}_{i}=P_{ii}+\sum_{{k}\in\textrm{N}(i)}
P_{ki}$,\\&&
$\quad\quad\tilde{\mu}_{i}=\tilde{P_{i}}^{-1}(P_{ii}\mu_{ii}+\sum_{k
\in \textrm{N}(i)} P_{ki}\mu_{ki})$, $\forall
i$\\&&\quad\quad\texttt{(under certain scheduling)}.\\&&
  $\checkmark$\quad\texttt{Compute the} $\textrm{N}(j)\ni i\rightarrow j$ \texttt{internal scalars} \\&& $\quad\quad P_{ij} = -A_{ij}^{2}/(\tilde{P}_{i}-P_{ji})$,\\
  &&$\quad\quad\mu_{ij}=(\tilde{P_{i}}\tilde{\mu_{i}}-P_{ji}\mu_{ji})/A_{ij}$.\\
  {\texttt{3.}} & {\emph{\texttt{Check:}}} & $\checkmark$\quad\texttt{If the internal scalars} $P_{ij}$ \texttt{and} $\mu_{ij}$ \texttt{did not}\\&&\quad\quad\texttt{converge (w.r.t. $\epsilon$),} \texttt{return to
    Step 2.}\\&&$\checkmark$\quad\texttt{Else, continue to Step 4.}\\
  {\texttt{4.}} & {\emph{\texttt{Infer:}}} & $\checkmark$\quad\texttt{Compute the marginal means}\\&&{\quad\quad$\mu_{i}=\big(P_{ii}\mu_{ii}+\sum_{k \in
\textrm{N}(i)}P_{ki}\mu_{ki}\big)/\big(P_{ii}+\sum_{{k}\in\textrm{N}(i)}
P_{ki}\big)=\tilde{\mu}_{i}$, $\forall i$.}\\&&$(\checkmark$\quad\texttt{Optionally compute the marginal precisions}\\&&\quad\quad$P_{i}=P_{ii}+\sum_{k\in\textrm{N}(i)}P_{ki}=\tilde{P}_{i}\quad)$\\
  {\texttt{5.}} & {\emph{\texttt{Solve:}}} & $\checkmark$\quad\texttt{Find the solution}\\&& {$\quad\quad x_{i}^{\ast}=\mu_{i}$, $\forall i$.}\\&&\\\hline
\end{tabular}}
\end{table}

Instead of sending a message composed of the pair of $\mu_{ij}$
and $P_{ij}$, a node can broadcast the aggregated sums \BEA
\tilde{P}_{i}&=&P_{ii}+\sum_{{k}\in\textrm{N}(i)}
P_{ki},\\\tilde{\mu}_{i}&=&\tilde{P}_{i}^{-1}(P_{ii}\mu_{ii}+\sum_{k
\in \textrm{N}(i)} P_{ki}\mu_{ki}). \EEA Consequently, each node
can retrieve locally the $P_{i\backslash j}$~ and
$\mu_{i\backslash j}$~ from the sums by means of a subtraction
\BEA P_{i\backslash j}&=&\tilde{P}_{i}-P_{ji},\\\mu_{i\backslash
j}&=&\tilde{\mu}_{i}-P_{i \backslash j}^{-1}P_{ji}\mu_{ji}.\EEA
The rest of the algorithm remains the same.

\section{Experimental Results}
\label{exp_results} \label{experimental} \label{emp_results} We
implemented our proposed algorithm using approximately 1,000 lines
of code in C. We implemented communication between the nodes using
the MPICH2 message passing interface ~\cite{MPI}. Each node was
responsible for $d$ data points out of the total $n$ data points
in the dataset.

Our implementation used synchronous communication rounds because
of MPI limitations. In Section~\ref{discussion} we further
elaborate on this issue.

Each node was assigned several examples from the input file. Then,
the kernel matrix $D$ was computed by the nodes in a distributed
fashion, so that each node computed the rows of the kernel matrix
related to its assigned data points. After computing the relevant
parts of the matrix $D$, the nodes weighted the diagonal of the
matrix $D$, as discussed in Section~\ref{diagonal}. Then, several
rounds of communication between the nodes were run. In each round,
using our optimization, a total of $n$ sums were calculated using
MPI\_Allreduce system call. Finally, each node output the solution
$x$, which was the mean of the input Gaussian that matched its own
data points. Each $x_i$ signified the weight of the data point $i$
for being chosen as a support vector.

To compare our algorithm performance, we used two
algorithms: Sequential SVM (SVMSeq) ~\cite{SVMSeq} and SVMlight
~\cite{SVMlight}. We used the SVMSeq implementation provided within the
IBM Parallel Machine Learning (PML) toolbox~\cite{IBM}. The PML implements the same algorithm
by Vijaykumar and Wu \cite{SVMSeq} that our GaBP solver is based on, but the implementation
in through a master-slave architecture as described in \cite{SVMSeqY}. SVMlight is a single computing node solver.

\begin{table}[t]
\begin{center}
\begin{tabular}{|l|l|l|l|c|c|c|}
\hline
Dataset                & Dimension & Train & Test & \multicolumn{3}{|c||}{\textsc{Error (\%)}} \\
\cline{5-7}
                         & & &                    & GaBP & Sequential    & SVMlight    \\
\hline
\texttt{Isolet}        & 617 & 6238 & 1559 & 7.06 & {\bf 5.84} & 49.97 \\
\texttt{Letter}        & 16 & 20000 &  & {\bf 2.06} & {\bf 2.06} & 2.3 \\
\texttt{Mushroom}      & 117 & 8124 & & 0.04 & 0.05 & {\bf 0.02}  \\
\texttt{Nursery}       & 25 & 12960 & & 4.16 & 5.29  & {\bf 0.02}  \\
\texttt{Pageblocks}   & 10 & 5473 & & 3.86 & 4.08  & {\bf 2.74}  \\
\texttt{Pen digits}    & 16 & 7494& 3498 & 1.66 & {\bf 1.37} & 1.57 \\
\texttt{Spambase}      & 57 & 4601 & & {\bf 16.3} & 16.5   & 6.57 \\
\hline
\end{tabular}
\caption{Error rates of the GaBP solver versus those of the parallel
sequential solver and SVMlight}
\label{err_run}
\end{center}
\end{table}

\begin{table}[t]
\begin{center}
\begin{tabular}{|l||c|c|}
\hline
Dataset                & \multicolumn{2}{|c|}{\textsc{Run times (sec)}}     \\
\cline{1-3}
                       & GaBP & Sequential     \\
\hline
\texttt{Isolet}        & 228     &  1328                  \\
\texttt{Letter}        & 468      &  601                    \\
\texttt{Mushroom}      & 226     &  176                   \\
\texttt{Nursery}       & 221     &  297                   \\
\texttt{Pageblocks}    & 26      &  37                    \\
\texttt{Pen digits}    & 45     &  155                   \\
\texttt{Spambase}      & 49      &  79                    \\
\hline
\end{tabular}
\caption{Running times (in seconds) of the GaBP solver (working in
a distributed environment) compared to that of the IBM parallel
solver} \label{time_run}
\end{center}
\end{table}

Table \ref{err_run} describes the seven datasets we used to
compare the algorithms and the classification accuracy obtained.
These computations were done using five processing nodes (3.5GHz
Intel Pentium machines, running the Linux operating system) for
each of the parallel solvers. All datasets were taken from the UCI
repository~\cite{UCI}. We used medium-sized datasets (up to 20,000
examples) so that run-times using SVMlight would not be
prohibitively high. All algorithms were run with an RBF kernel.
The parameters of the algorithm (kernel width and
misclassification cost) were optimized using
 a line-search algorithm, as detailed in~\cite{RifkinK04}.

Note that SVMlight is a single node solver, which we use mainly as a comparison
for the accuracy in classification.

Using the Friedman test \cite{Demsar06}, we did not detect any statistically
significant difference between the performance of the algorithms with regards
 to accuracy ($p<0.10^{-3}$).

\begin{figure}[t]
\begin{center}
  \includegraphics[width=340pt]{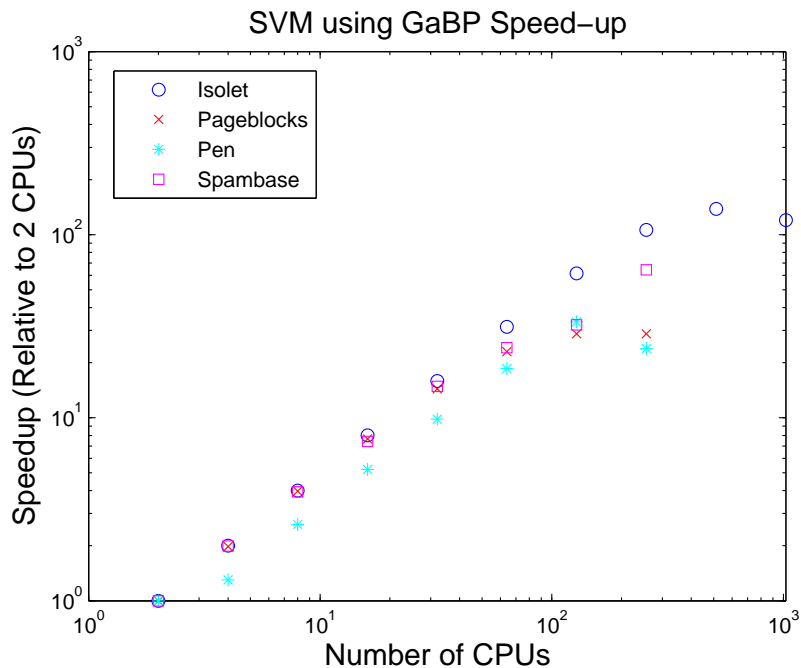}\\
  \caption{Speedup of the GaBP algorithm vs. 2 CPUS}\label{speedup}
\end{center}
\end{figure}

Figure~\ref{speedup} shows the speedup results of the algorithm when running the GaBP algorithm on a Blue Gene supercomputer.
The speedup with $N$ nodes is computed as the run time of the algorithm on a single node,
divided by the run time using $N$ nodes. Obviously, it is desirable to obtain linear speedup, i.e.,
doubling computational power halves the processing time, but this is limited by the communication load and
by parts of the algorithm that cannot be parallelized.
Since Blue Gene is currently limited to 0.5 GB of memory at each node, most datasets could not
be run on a single node. We therefore show speedup compared to two nodes.
As the figure shows, in most cases we get a linear speedup up to 256
CPUs, which means that the running time is linearly
proportional to one over the number of used CPUs. When using 512 - 1024
CPUs, the communication overhead reduces the efficiency of the
parallel computation. We identified this problem as an area for future
research into optimizing the performance for larger scale grids.

We also tested the ability to build classifiers for larger
datasets. Table \ref{bg_times} shows the run times of the GaBP
algorithm using 1024 CPUs on two larger datasets, both from the
UCI repository. This demonstrates the ability of the algorithm to
process very large datasets in a reasonably short amount of time.
We compare our running time to state-of-the-art parallel
decomposition method by Zanni \etal ~\cite{Zanni} and Hazan \etal.
Using the MNIST dataset we where considerably slower by a factor
of two, but in the larger Covertype dataset we have a superior
performance. Note that it is hard to compare running times since
the machines used for experimentation are different. Zanni used 16
Pentium IV machines with 16Gb memory, Hazan used 10 Pentium IV
machines with 4Gb memory while we used a larger number of weaker
Pentium IV machines with 400Mb of memory. Furthermore, in the
Covertype dataset we used only 150,000 data points while Zanni and
Hazan used the full dataset which is twice larger.

\begin{table}[t]
\begin{center}
\begin{tabular}{|l|l|l|c|c|c|}
\hline
Dataset                   & Dim       & Num of examples    & Run time GaBP (sec) & Run time ~\cite{Zanni} (sec) & Run time~\cite{Hazan}\\
\hline
\texttt{Covertype}        & 54                      & 150,000/300,000        & {\bf 468}   & 24365 & 16742\\
\texttt{MNIST}            & 784                     & 60,000         & 756   & 359 &  {\bf 18}\\
\hline
\end{tabular}
\caption{Running times of the GaBP solver for large data sets
using 1024 CPUs on an IBM Blue Gene supercomputer. Running time
results are compared to two state-of-the-art solvers:
~\cite{Zanni} and~\cite{Hazan}.} \label{bg_times}
\end{center}
\end{table}

\section{Discussion}
\label{discussion}

In this paper we demonstrated the application of the Gaussian Belief Propagation to the solution of SVM problems.
Our experiments demonstrate the usefulness of this solver, being both accurate and scalable.

We implemented our algorithm using a synchronous communication model mainly because MPICH2 does not support
asynchronous communication. While synchronous communication is the mode of choice for supercomputers such as Blue Gene,
in many cases such as heterogeneous grid environments, asynchronous communication will be preferred.
We believe that the next challenging goal will be to implement the proposed algorithm in asynchronous settings, where
algorithm rounds will no longer be synchronized.

Our initial experiments with very large sparse kernel matrices (millions of data points) show that asynchronous settings converge
faster. Recent work by Koller~\cite{RBP} supports this claim by showing that in many cases the BP algorithm converges faster in
asynchronous settings.

Another challenging task would involve scaling to data sets of millions of data points. Currently the full
kernel matrix is computed by the nodes. While this is effective for problems with many support vectors \cite{SVMSeqY}, it
is not required in many problems which are either easily separable or else where the classification error is less
important compared to the time required to learn the mode. Thus, solvers scaling to much larger datasets may have to
diverge from the current strategy of computing the full kernel matrix and instead sparsify the
kernel matrix as is commonly done in single node solvers.

Finally, it remains an open question whether SVMs can be solved
efficiently in Peer-to-Peer environments, where each node can
(efficiently) obtain data from only several close peers. Future
work will be required in order to verify how the GaBP algorithm
performs in such an environment, where only partial segments of
the kernel matrix can be computed by each node.


\begin{thebibliography}{10}

\bibitem{Weiss}
Y. Weiss and W. T. Freeman. Correctness of belief propagation in
Gaussian graphical models of arbitrary topology. In NIPS-12, 1999

\bibitem{J}
J.K. Johnson. Walk-summable Gauss-Markov random fields. Technical
Report, February 2002. (Corrected, November 2005).

\bibitem{RBP}
G. Elidan and I. McGraw and D. Koller, Residual Belief
Propagation: Informed Scheduling for Asynchronous Message Passing,
Proceedings of the Twenty-second Conference on Uncertainty in AI
(UAI), Boston, Massachussetts, 2006

\bibitem{WS}
J.K. Johnson, D.M. Malioutov, A.S. Willsky. Walk-sum
interpretation and analysis of Gaussian belief propagation, In
Advances in Neural Information Processing Systems, vol. 18, pp.
579-586, 2006.

\bibitem{JMLR}
D.M. Malioutov, J.K. Johnson, A.S. Willsky. Walk-sums and belief
propagation in Gaussian graphical models, Journal of Machine
Learning Research, vol. 7, pp. 2031-2064, October 2006.

\bibitem{SVM}
Nello Cristianini and John Shawe-Taylor. An Introduction to Support Vector Machines and Other
Kernel-based Learning Methods. Cambridge University Press, 2000. ISBN 0-521-78019-5.

\bibitem{SVMSeq}
Sethu Vijayakumar and Si Wu (1999), Sequential Support Vector
Classifiers and Regression. Proc. International Conference on Soft
Computing (SOCO'99), Genoa, Italy, pp.610-619.

\bibitem{SVMSeqY}
Elad Yom-Tov (2007), A distributed sequential solver for large scale SVMs. In: O. Chapelle, D. DeCoste, J. Weston, L. Bottou: Large scale kernel machines. MIT Press, pp. 141-156.

\bibitem{SS2002}
B. Sch{\"o}lkopf and A. J. Smola. Learning with kernels: Support
vector machines, regularization, optimization, and beyond. MIT
Press, Cambridge, MA, USA, 2002.

\bibitem{IBM}
{\tt http://www.alphaworks.ibm.com/tech/pml}

\bibitem{UCI}
Catherine~L. Blake, Eamonn~J. Keogh, and Christopher~J. Merz.
{UCI} repository of machine learning databases, 1998.
URL \url{http://www.ics.uci.edu/$\sim$mlearn/MLRepository.html}.

\bibitem{RifkinK04}
R.~Rifkin and A.~Klautau.
In defense of {One-vs-All} classification.
Journal of Machine Learning Research, 5:101--141, 2004.

\bibitem{Demsar06}
J.~Dem\u{s}ar.
Statistical comparisons of classifiers over multiple data sets.
Journal of Machine Learning Research, 7:1--30, 2006.

\bibitem{ZZ2003}
G. Zanghirati and L. Zanni. A parallel solver for large quadratic
programs in training support vector machines. Parallel computing,
29:535-551, 2003.

\bibitem{CBB2002}
R. Collobert, S. Bengio, and Y. Bengio. A parallel mixture of svms
for very large scale problems. In Advances in Neural Information
Processing Systems. MIT Press, 2002.

\bibitem{SVMlight}
T.~Joachims.
Making large-scale svm learning practical.
In "B. Sch{\"o}lkopf, C. Burges, A. Smola" (Editors),
{\em Advances in Kernel Methods - Support Vector Learning},

\bibitem{GCBD2004}
H.~P. Graf, E.~Cosatto, L.~Bottou, I.~Durdanovic, and V.~Vapnik.
Parallel support vector machines: The cascade svm.
In {\em Advances in Neural Information Processing Systems}, 2004.

\bibitem{ISIT2}
D.~Bickson, O.~Shental, P.~H. Siegel, J.~K. Wolf, and D.~Dolev.
Gaussian belief propagation based multiuser detection. In {\em
IEEE Int. Symp. on Inform. Theory (ISIT)}, Toronto, Canada, July
2008, to appear.

\bibitem{ISIT1}
O. Shental, D. Bickson, P.~H. Siegel, J.~K. Wolf and D. Dolev.
Gaussian belief propagation solver for systems of linear
equations.In {\em IEEE Int. Symp. on Inform. Theory (ISIT)},
Toronto, Canada, July 2008, to appear.

\bibitem{NIPS-workshop}
D.Bickson, D. Dolev and E. Yom-Tov, Solving Large Scale Kernel
Ridge Regression using A Gaussian Belief Propagation Solver {\em
in NIPS Workshop on  Efficient Machine Learning}, Canada, 2007.

\bibitem{Allerton}
D.~Bickson, O.~Shental, P.~H. Siegel, J.~K. Wolf, and D.~Dolev.
Linear detection via belief propagation, in {\em 45th Allerton
Conf. on Communications, Control and Computing}, Monticello, IL,
USA, Sept. 2007.


\bibitem{Zanni}
L. Zanni, T. Serafini and Gaetano Zanghirati. Parallel Software
for Training Large Scale Support Vector Machines on Multiprocessor
Systems. In {\em proc of Journal of Machine Learning Research,
Vol. 7} (July 2006), pp. 1467-1492.

\bibitem{Hazan}
T. Hazan, A. Man and A. Shashua.  A Parallel Decomposition Solver
for SVM: Distributed Dual Ascent using Fenchel Duality. In {\em
Conference on Computer Vision and Pattern Recognition (CVPR)},
Anchorage, June 2008, to appear.

\bibitem{MPI}
MPI message passing interface. {\tt
http://www-unix.mcs.anl.gov/mpi/mpich/}

\end{thebibliography}
\end{document}